\documentclass[lettersize,journal]{IEEEtran}
\usepackage[colorlinks=true,linkcolor=black,urlcolor=black,citecolor=black]{hyperref}
\usepackage{amsmath,amsfonts}
\usepackage{algorithmic}
\usepackage{algorithm}
\usepackage{array}
\usepackage[caption=false,font=normalsize,labelfont=sf,textfont=sf]{subfig}
\usepackage{textcomp}
\usepackage{stfloats}
\usepackage{url}
\usepackage{verbatim}
\usepackage{graphicx}
\usepackage{cite}
\usepackage{booktabs}
\usepackage{tabularx}
\usepackage{soul}
\usepackage{color, xcolor}
\usepackage{tabularx} 
\begin{document}

\title{Surface Constraint Policy for Learning \\
Surface-Constrained and Dynamically Feasible Robot Skills}

\author{Shuai Ke, Jiexin Zhang,~\IEEEmembership{Member,~IEEE}, Huan Zhao,~\IEEEmembership{Member,~IEEE}, 
Zhiao Wei, Yikun Guo, Jie Pan,\\ and Han Ding,~\IEEEmembership{Senior Member,~IEEE}%
\thanks{This work was supported by the National Natural Science Foundation of China under Grants U24A20130, 52505016, and 52188102. (Corresponding author: Jiexin Zhang.)}%
\thanks{The authors are with the State Key Laboratory of Intelligent Manufacturing Equipment and Technology, Huazhong University of Science and Technology, Wuhan 430074, China.(e-mail: keshuai@hust.edu.cn; zhangjiexin@hust.edu.cn; huanzhao@hust.edu.cn; zhiao\_wei@hust.edu.cn; guoyikun@hust.edu.cn; panjie@hust-wuxi.com; dinghan@hust.edu.cn.)}
}
\markboth{Journal of \LaTeX\ Class Files}%
{Shell \MakeLowercase{\textit{et al.}}: A Sample Article Using IEEEtran.cls for IEEE Journals}

\maketitle

\begin{abstract}
Diffusion-based imitation learning methods have driven rapid progress in robot dexterous manipulation tasks. However, they have limitations when applied to tasks that involve complex free-form surface constraints because of their lack of explicit surface geometry constraint modeling and the dynamic feasibility issue, resulting in stochastic action generation that fails to achieve reliable surface alignment and maintain stable contact. To address these limitations, we propose a novel surface constraint policy (SCP) for generating robot actions that satisfy free-form surface constraints on the basis of human demonstrations and real-time visual observations. First, the surface geometry constraint is encoded using a two-dimensional weighted Gaussian kernel function that is derived from demonstrations. Building on the encoded surface geometry constraints, the diffusion-based policy is used to infer task-level action intentions from multimodal sensory inputs, including visual observations and robot state feedback. These intentions are further transformed into surface-constrained dynamic movement primitives (DMPs) through a similarity-based action mapping method, thereby enabling smooth and compliant motion execution. The SCP achieves generation of structured surface geometric intent and dynamically admissible actions. The proposed method is validated on multiple surface manipulation tasks and compared with existing techniques. The experimental results demonstrate superior task success rates and contact stability under surface constraints.
\end{abstract}


\section{Introduction}
\IEEEPARstart{I}{mitation} learning for robots has rapidly advanced in recent years. Early approaches relied primarily on explicit modeling of demonstration trajectories~\cite{xing2023,liu2022}, such as Gaussian mixture models (GMMs)~\cite{Ma2020}, hidden Markov models (HMMs)~\cite{zhang2023}, and kernelized movement primitives (KMPs), which reproduce demonstrated behaviors by fitting trajectory distributions in structured spaces. Inverse reinforcement learning (IRL)~\cite{adams2022,zhang2021} was subsequently developed to infer latent reward functions from demonstrations, thereby reducing the dependence on manually designed and structured state--action mappings~\cite{urain2024}. 
While these approaches have alleviated the dependence on explicitly designed structured state–action mappings, their performance remains constrained when handling rich multimodal sensory inputs and achieving robust generalization in complex robotic manipulation tasks~\cite{li2024,zhang2025}.

Generative-model-based imitation learning approaches have attracted increasing attention and are emerging as a key direction for enhancing robot skill generalization and task adaptability. For example, generative adversarial networks~\cite{aggarwal2021}, action chunking with transformers (ACT)~\cite{zhao2023}, and diffusion-based approaches~\cite{li2024,chi2023,zhou2024} involve attempting to directly map sensory inputs to complete action trajectories in an end-to-end fashion. These approaches demonstrate strong robustness and diversity~\cite{chi2023} and achieve improved policy stability and generalization across task variations and environmental perturbations while avoiding the need for complex reward functions or handcrafted features. They have achieved notable success in tasks such as deformable object manipulation~\cite{scheikl2024}, pouring~\cite{liu2024rdt,yu2025}, and assembly~\cite{ze2024}.

\begin{figure}[t]
    \centering
    \includegraphics[scale=0.07, angle=0, origin=c]{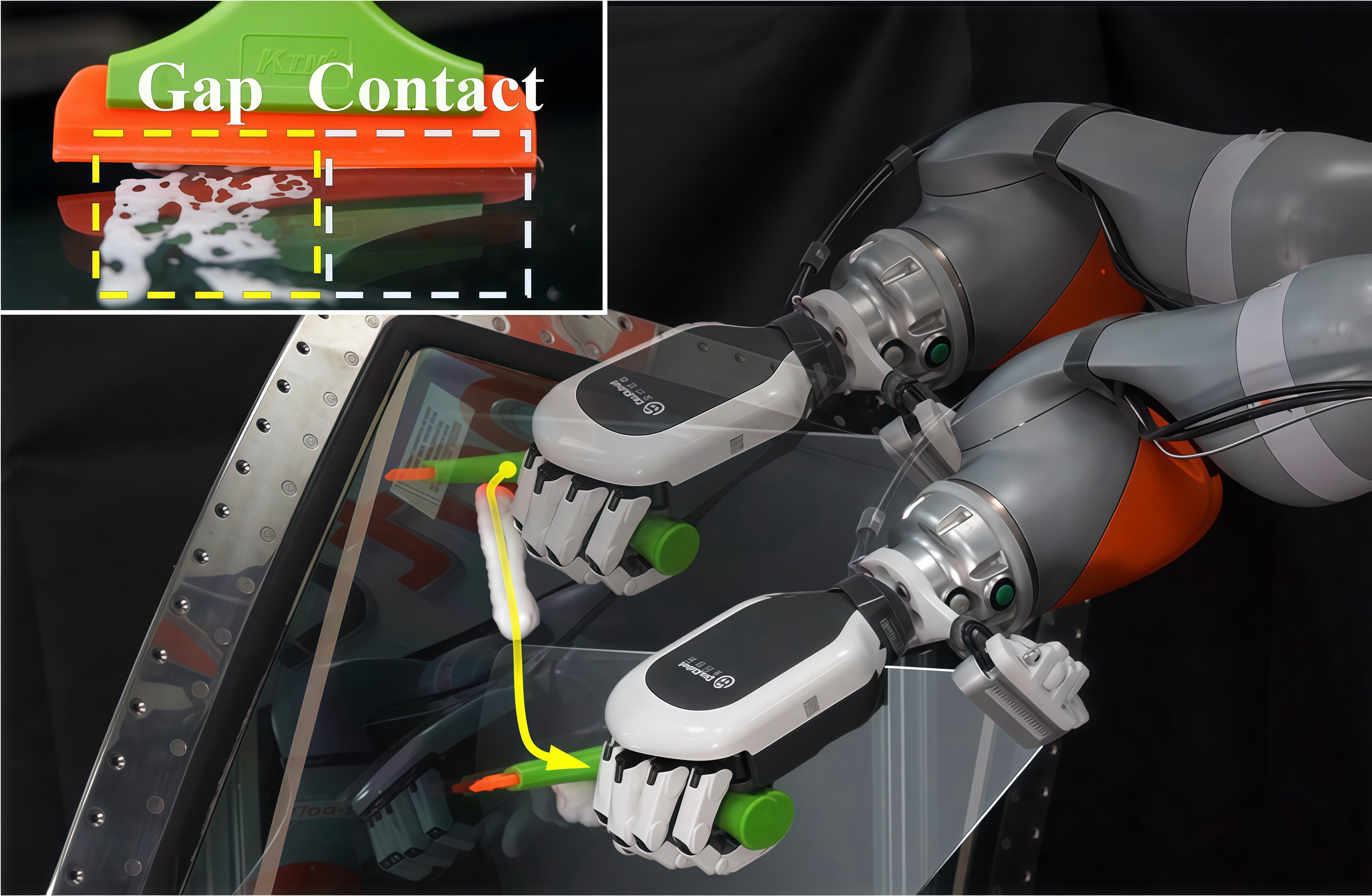}
    \caption{Example of a surface-constrained robotic task: Aircraft windscreen cleaning. The robot is required to grasp a cleaning tool and perform wiping along the glass surface while maintaining stable contact at the correct angle to remove surface stains. Areas with contact gaps that are caused by incorrect tool angles or unstable contact may lead to residual stains.}
    \label{fig1}
\end{figure}

However, generative-model-based imitation learning approaches still face challenges in handling tasks that involve geometric constraints on complex surfaces, such as wiping~\cite{liu2025}, polishing~\cite{wang2025}, and ultrasound scanning~\cite{li2021,huang2018}. Fig.\ref{fig1} illustrates a typical surface-constrained task. These tasks impose dual requirements on the robot, namely, that its contact posture must follow the geometric variations of the target surface, while the contact process must remain continuous and stable~\cite{ke2025}. Current mainstream approaches focus primarily on global motion planning or coarse-grained task-level control~\cite{chi2023,liu2024rdt}, while lacking explicit surface geometry constraint modeling and dynamic feasibility considerations. The lack of explicit surface geometry constraint modeling leads to increased surface-alignment errors, while insufficient consideration of dynamic feasibility results in large motion accelerations and irregular trajectory curvature. In this work, dynamic feasibility refers to the smooth and stable executability of the generated trajectories on a real robot. The combined effects of these issues often cause loss of contact or surface damage during task execution, ultimately degrading task completion efficiency and reducing success rates. In various studies, constraints have been introduced during training~\cite{liu2021,bastek2024,giannone2023,christopher2024}; however, effective generalization to freeform surface tasks remains challenging. In robot manipulation tasks involving surface geometry constraints, existing methods still face challenges in fine-grained geometric action intent generation and dynamically feasible execution.

\subsection{Related Work}
To comprehensively understand the capabilities and limitations of existing methods in these aspects, we further review imitation learning approaches involving geometric constraints before presenting our proposed method and contributions. The related work is analyzed from the perspectives of precision action intent generation and dynamic behavior execution.

\subsubsection{Precision Action Intent Generation}
Currently, action generation strategies that are based on generative models struggle to directly infer robot behaviors subject to surface geometry constraints. In one line of research, researchers have sought to improve the granularity and accuracy of inferred action intent by enhancing visual input quality. For instance, Ze et al.~\cite{ze2024} proposed a 3D diffusion policy to incorporate 3D point cloud information for fine-grained behavior generation, thereby significantly improving manipulation performance. Similarly, Chi et al.~\cite{chi2024umi} developed a handheld gripper that records pose data and improves monocular vision with mirror extensions during demonstrations to support precise manipulation learning. Despite these advances, owing to the absence of explicit constraints on the generated output actions, ensuring geometric consistency remains challenging. 

To address this, in various works, researchers have attempted to introduce constraints directly into generative models. Bastek et al.~\cite{bastek2024} proposed physics-informed diffusion models, which apply physics-based loss functions to encourage physically plausible samples. Nevertheless, such methods cannot strictly enforce constraint satisfaction. Giannone et al.~\cite{giannone2023} further introduced diffusion optimization models and incorporated a trajectory-alignment mechanism to optimize behavior during sampling. However, this approach fails to account for data uncertainty during optimization, thus often resulting in behavior that significantly deviates from the original data distribution. 

Other studies have ensured interaction with manipulated objects by reasoning about contact forces. Notably, Aburub et al. proposed the DIPCOM~\cite{aburub2024}, a diffusion-based policy learning approach designed for compliant, contact-rich manipulation. Distinct from traditional methods that solely plan trajectories or forces, DIPCOM leverages the powerful multi-modal distribution modeling capabilities of diffusion models to simultaneously predict end-effector poses and dynamically adjust arm stiffness. This enables the implicit learning of adaptive compliance strategies directly from expert demonstrations. In contrast, Hou et al. introduced the adaptive compliance policy (ACP)~\cite{hou2024}, which jointly predicts trajectories and plans contact forces and then adjusts robot motion accordingly. However, such methods may face limitations when applied to tasks that involve complex tools, where fine-grained contact modeling and tool--object interaction become more challenging.

\subsubsection{Dynamic Behavior Execution}
Current generative strategies often exhibit poor execution performance when deployed on real robotic systems, primarily because of the lack of explicit modeling of kinematic and dynamic constraints, which results in unstable or infeasible behaviors during task execution. Prior studies have highlighted a fundamental contradiction between the inherent stochasticity of generative models and the precise dynamics that are required for robotic control~\cite{bouvier2025}. To address this, one line of research have focused on improving the smoothness and controllability of generated trajectories. For example, Xirui Shi et al. proposed FRMD~\cite{shi2025} to achieve fast and smooth single-step trajectory generation on the basis of diffusion policies and demonstrated its effectiveness in the PlugCharger-v1 task. Similarly, Scheikl et al. introduced MPD~\cite{scheikl2024}, which leverages ProDMPs~\cite{li2023prodmp} to generate smooth action sequences and improves both the feasibility and compliance of trajectories in deformable object manipulation tasks. These works suggest that generating smoother trajectories can significantly improve task success rates in practice. 

In addition, several approaches introduce physical constraint modeling directly into the diffusion generation process. Christopher et al.~\cite{christopher2024} proposed integrating constraint projection into the denoising process to ensure that the generated actions strictly satisfy task-specific constraints. Although this class of methods theoretically guarantees physical feasibility, they often require repeated projection steps when the initial solution deviates significantly from the feasible set, thereby resulting in substantial computational overhead and limited scalability for long-horizon tasks. Furthermore, Bouvier et al. introduced DDAT~\cite{bouvier2025}, which is a framework in which diffusion models are used to generate provably admissible trajectories for black-box robotic systems. Overall, current generative strategies frequently suffer from unstable execution and infeasible trajectories because of inadequate modeling of kinematic and dynamic constraints. For instance, trajectory discontinuities are common in generated action sequences and severely affect the quality and robustness of robot task execution.

In summary, current diffusion-based methods for free-form surface-constrained motion generation exhibit two key limitations. First, the inferred robot action intents are neither sufficiently precise nor compliant with the surface geometry constraints that are required by the task, which causes the generated motions to lack structural alignment with the target surface. Second, existing methods have difficulty converting the inferred intent into dynamically feasible behaviors, which leads to nonsmooth trajectories, abrupt accelerations, and unstable execution. Therefore, for tasks involving complex surface geometry constraints, designing generative strategies that can both encode structured surface-geometry intent and ensure dynamically feasible control remains an open challenge.

\begin{figure*}[!t]
\centering
\subfloat{\includegraphics[scale=0.3]{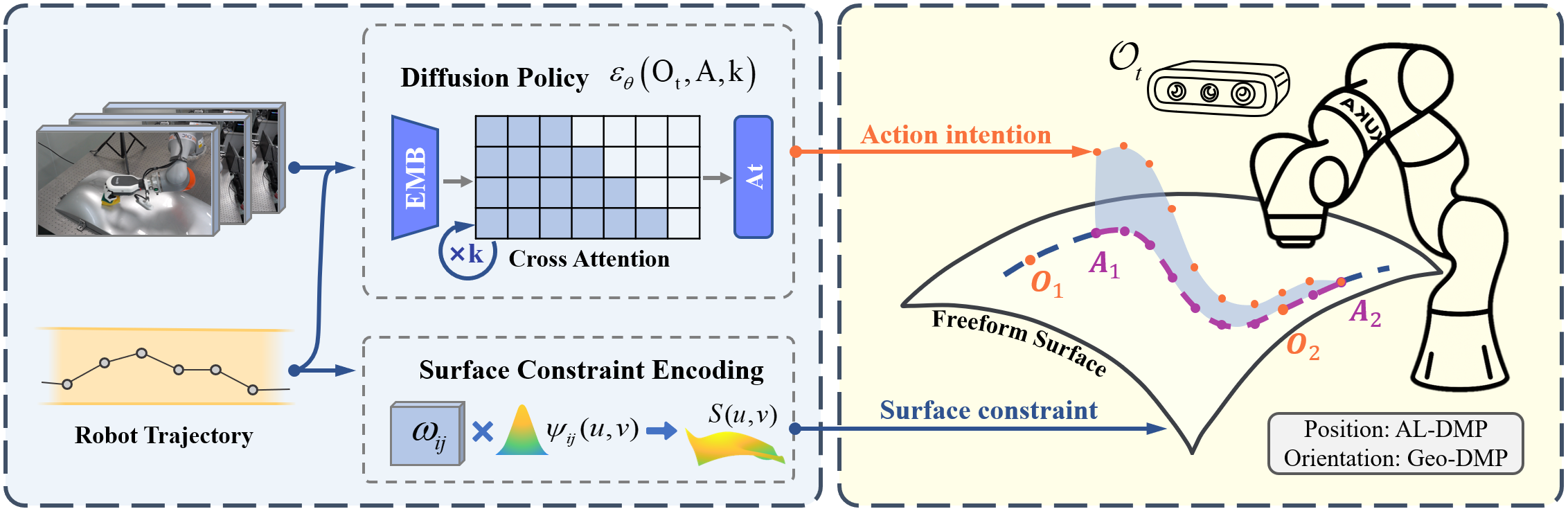}}%
\hfil
\caption{Pipeline for the Surface Constraint Policy. The system generates robot actions with surface geometry constraints on the basis of prior human demonstrations and real-time visual observations. Left: Using locally weighted regression (LWR) on demonstrated trajectories, a 2D weighted Gaussian kernel weight matrix $w_{ij}$ is computed to encode the surface $S(u, v)$ (Section~II-B). The diffusion policy generates action intention $A_t$ through a cross-attention mechanism (Section~II-C). Right: The action intention that is inferred at moment $O_1$ (orange dots) is mapped onto a surface-constrained target action (purple dots) via the similarity mapping method (blue area, Section~III-A). Finally, position and orientation primitives that are modulated by the geodesic phase are used to control the robot and execute surface-constrained actions (Section~III-B).}
\label{fig2}
\end{figure*}

\subsection{Main Contributions}
Motivated by the above limitations, we propose a surface constraint policy, which is a novel policy for enabling robots to perform tasks subject to free-form surface constraints. The SCP can be used to encode surface-constrained task intent from human demonstrations and generate corresponding constraint-compliant robotic actions. The pipeline for the SCP is illustrated in Fig.\ref{fig2}.

We systematically evaluate the proposed method on a wiping task and compare it with several state-of-the-art baselines. The experimental results show that our method significantly improves the success rate of surface-constrained tasks and effectively enhances the accuracy of action intent generation as well as the smoothness and stability of robot actions. The contributions of this work can be summarized as follows:

\begin{enumerate}
   \item An SCP is proposed as a policy learning framework for generating actions under surface geometry constraints, producing fine-grained and geometry-consistent intents.

    \item A similarity mapping method is proposed for transforming policy intentions into dynamically feasible robot actions, thereby increasing action smoothness and stability.

    \item The proposed method is evaluated on three tasks with free-form surface constraints. Results demonstrate its effectiveness and generalizability in manipulation.
\end{enumerate}

The remainder of the paper is organized as follows. Section II introduces the action generation framework that integrates surface DMPs with diffusion-based strategies. Section III presents the surface-constrained motion mapping approach, which is based on the inference results of the diffusion model. Section IV validates the effectiveness of the proposed method through experiments. Finally, Section V concludes the paper.

\section{Surface Constraint Policy Framework}
In this section, we present the SCP framework, which integrates diffusion-based action intent with explicit surface geometry constraints to generate robot motions that satisfy both surface-constrained behavior and dynamic feasibility. Section II-A presents the overall pipeline of SCP, Section II-B details the surface information encoding method, and Section II-C describes the action intent inference process, which is based on the diffusion policy.

\subsection{Pipeline for the Surface Constraint Policy}
In this section, the pipeline of the proposed SCP, which is a generative action framework that integrates surface DMPs with a diffusion policy for generating robot actions subject to surface geometric constraints, is introduced. Overall, the SCP generates constrained actions on the basis of the demonstration trajectory set
\begin{align}
\mathcal{T} = \left\{ \tau_k \mid \tau_k = \left\{ \left( \mathbf{p}_i^{(k)}, \mathbf{q}_i^{(k)}, \mathbf{f}_i^{(k)} \right) \right\}_{i=1}^{N_k} \right\}
\end{align}
where $\mathbf{p}_i^{(k)} \in \mathbb{R}^3$ denotes the cartesian position, $\mathbf{q}_i^{(k)} \in \mathbb{S}^3$ represents the orientation in quaternion form, and $\mathbf{f}_i^{(k)} \in \mathbb{R}$  represents the normal contact force of the $i$-th point in the $k$-th trajectory $\tau_k$ (which contains $N_k$ points). where $\mathbb{S}^3$ denotes the unit quaternion manifold, i.e., the set of quaternions with unit norm. The corresponding image observations $\mathcal{O}_t \in \mathbb{R}^{H \times W \times 3}$ are also collected during demonstrations. 

Specifically, during the demonstration phase, a human operator teleoperates the robot to perform a task. An observation camera records an image sequence $\mathcal{O}_t$, while the robot returns its current end-effector pose $\mathbf{P}_i = (\mathbf{p}_i, \mathbf{q}_i)$, where $\mathbf{p}_i \in \mathbb{R}^3$ and $\mathbf{q}_i \in \mathbb{S}^3$, and joint configuration $\mathbf{J}_i \in \mathbb{R}^7$. These data are temporally synchronized and stored.

Next, for surface geometry encoding, the end-effector poses $\mathbf{P}_i = (\mathbf{p}_i, \mathbf{q}_i)$ that are collected during contact with the surface form a dataset $\mathcal{D} = \{ (\mathbf{p}_i, \mathbf{q}_i) \}_{i=1}^{M}$. This dataset is passed to a surface encoding module, which uses a 2D weighted Gaussian kernel function to construct a surface state representation $S(u,v)$, which provides accurate geometric priors for subsequent policy learning.

The observation at time $t$ is defined as follows:
\begin{align}
\mathbf{O}_t = \left\{ \mathcal{O}_t \right\} \cup \left\{ \mathbf{P}_t \right\} \cup \left\{ \mathbf{J}_t \right\}
\end{align}
where $\mathcal{O}_t \in \mathbb{R}^{H \times W \times 3}$ represents the RGB image, $\mathbf{P}_t$ represents the end-effector pose, and $\mathbf{J}_t$ represents the joint configuration. \textit{Note: The calligraphic symbol $\mathcal{O}_t$ denotes the RGB image sequence, whereas the regular symbol $\mathbf{O}_t$ represents the full observation, including the pose and joint states.} These observations are used to train a task policy $\varepsilon_\theta(\mathbf{O}, \mathbf{A}, k)$ that maps perceptions to actions under demonstration supervision.

The policy inference at time $t$ outputs an action intention represented as a predicted trajectory
\begin{align}
\mathbf{A}_t = \left\{ \left( \mathbf{p}_t^{\text{pred}}, \mathbf{q}_t^{\text{pred}} \right) \right\}_{t = t_0}^{t_0 + N}.
\end{align}
These actions are then mapped onto the encoded surface using the proposed motion mapping method to generate the parameters for the DMPs.

Finally, a DMP-based controller is used to execute the robot trajectory in real time. In each inference cycle, the initial state $(\mathbf{x}_0, \dot{\mathbf{x}}_0)$ is set using the current position $\mathbf{p}_{\text{current}}$ and velocity $\dot{\mathbf{p}}_{\text{current}}$ of the robot, and the resulting motion is ensured to adhere to surface constraints with good continuity and robustness. This pipeline enables the efficient and stable execution of contact-rich tasks such as wiping and polishing.

\subsection{Surface Constraint Encoding}

To ensure that the generated actions satisfy the geometric constraints of the task surface, explicitly encoding the geometric features of the freeform surface to provide geometric priors for robot motion generation is necessary. In this section, a method for encoding surfaces using two-dimensional weighted Gaussian kernel functions and estimating orientations on the basis of intrinsic Gaussian clustering methods is presented~\cite{ke2025}. First, we parameterize the position $\mathbf{p}_i \in \mathbb{R}^3$ of the trajectory point set $D$ from demonstrations using $uv$ coordinates in a local reference frame. The corresponding 2D parametric coordinates $(u_i, v_i)$ are defined as follows:
\begin{equation}
(u_i, v_i) = \left( \frac{\mathbf{e}_x^\mathrm{T} (\mathbf{p}_i - \mathbf{o})}{s_x}, \frac{\mathbf{e}_y^\mathrm{T} (\mathbf{p}_i - \mathbf{o})}{s_y} \right)
\label{eq:uvparam}
\end{equation}
where $\mathbf{e}_x$ and $\mathbf{e}_y$ denote the basis vectors of the local coordinate system, $\mathbf{o}$ denotes the origin, and $s_x$ and $s_y$ represent scaling factors that are used to normalize the coordinate scales in different directions. We then construct a continuous surface state function $S(u, v)$ using a weighted Gaussian kernel function to encode the geometric shape of the freeform surface. The function is defined as follows:
\begin{equation}
S(u, v) = \frac{\sum\limits_{i=1}^{N} \sum\limits_{j=1}^{M} \omega_{ij} \psi_{ij}(u, v)}{\sum\limits_{i=1}^{N} \sum\limits_{j=1}^{M} \psi_{ij}(u, v)}
\label{eq:surface_state}
\end{equation}
where the kernel function $\psi_{ij}(u, v)$ is defined as a two-dimensional Gaussian
\begin{equation}
\psi_{ij}(u, v) = \exp\left( -h_{ij} \left\| (u, v) - \mathbf{c}_{ij} \right\|^2 \right),
\label{eq:gaussian_kernel}
\end{equation}

where $\mathbf{c}_{ij} \in \mathbb{R}^2$ represents the center of the Gaussian kernel, $h_{ij}$ represents the bandwidth parameter, and the weights $\omega_{ij}$ can be learned using locally weighted regression (LWR).

\subsection{Diffusion-Based Motion Generation}

We adopt a diffusion policy to infer the action intention of the robot. At each control timestep, the policy is used to generate a sequence of future actions conditioned on the most recent observations. Specifically, we formulate a visuomotor policy as a conditional denoising diffusion probabilistic model (DDPM). On the basis of the current RGB image sequence $\mathcal{O}_t \subset \mathbb{R}^{H \times W \times 3}$ captured by the camera, the end-effector pose $P_t = (p_t, q_t)$, and the joint angles $\mathbf{J}_t$, the model predicts the upcoming robot action intention $\mathbf{A}_t$.

For the multimodal and multisource observation inputs, we follow Chi et al.~\cite{chi2023} and use an untrained ResNet-18 to encode the RGB images. The proprioceptive state (i.e., time series data of $P_t$ and $J_t$) of the robot is encoded via a multilayer perceptron (MLP) or into latent features. These visual and state features are then concatenated and passed to a transformer-based diffusion model: Each position in the noisy action sequence is treated as a token, and denoising gradients are learned through the transformer decoder blocks. More concretely, both the noisy action sequence (which includes future poses at all timesteps) and the current diffusion step index $k$ are embedded as input tokens, and the observation embeddings are mapped via a shared MLP and provided as conditioning inputs to the transformer. In this way, the visual and proprioceptive embeddings provide guidance throughout the transformer decoding process for action generation.

The denoising iteration follows the standard DDPM procedure~\cite{Ho2020}. Starting from an initial action sequence $A_t^k$ that is sampled from a multidimensional Gaussian distribution, the model performs iterative denoising to gradually reduce noise and obtain the target sequence. At diffusion step $k$, the update is defined as follows:

\begin{equation}
A_t^{k-1} = \alpha \left( A_t^k - \gamma \, \epsilon_\theta \left( A_t^k, \mathcal{O}_t, k \right) \right) + \mathcal{N}(0, \sigma^2 I),
\label{eq:ddpm-denoise}
\end{equation}

where $\epsilon_\theta$ represents a parameterized noise prediction network that estimates the noise that is present in the current action. This step can be viewed as a stochastic gradient descent with noise, where the denoising direction is guided by the gradient estimated via $\epsilon_\theta$.

During training, a random diffusion step $k$ is selected for each training sample. Gaussian noise $\varepsilon^k$ is added to the ground-truth action sequence $A_t^0$, and the network is trained to predict this noise. The loss function is defined as the mean squared error (MSE) between the predicted and actual noise

\begin{equation}
\mathcal{L} = \text{MSE} \left( \varepsilon^k, \epsilon_\theta \left( \mathcal{O}_t, A_t^0 + \varepsilon^k, k \right) \right)
\label{eq:ddpm-loss}
\end{equation}
where $\mathcal{O}_t$ denotes the complete observation input at the current timestep. By minimizing this loss, the network learns to model the conditional distribution of actions given observations and implicitly approximates the distribution of demonstrated behavior.

\section{Similarity Mapping Method from Policy to Motion}
\begin{figure}[t]
    \centering
    \includegraphics[scale=0.31, angle=0, origin=c]{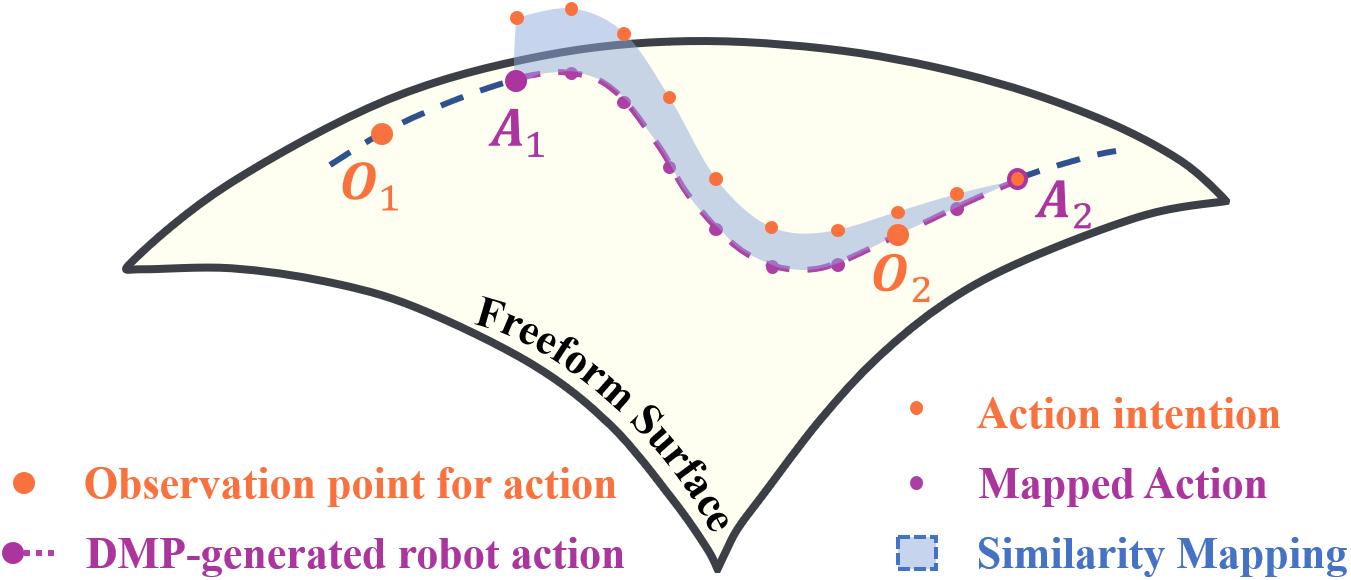}
    \caption{Illustration of the action intention mapping method. Joint optimization ensures surface constraint compliance and trajectory consistency by minimizing the Euclidean distance and the directional entropy KL divergence. To avoid motion lag, the next action segment is estimated before the previous segment finishes.}
    \label{fig3}
\end{figure}

Directly executing action intentions that are inferred from diffusion policies as trajectory points often fails to produce dynamically feasible and surface-geometry-constrained robot actions. To address issues such as motion discontinuity and poor stability that are caused by insufficient consideration of dynamics and surface geometry constraints, we propose a similarity-based mapping approach that transforms high-level policy intentions into surface-constrained DMPs. An illustration of the similarity mapping method is shown in Fig. \ref{fig3}. Section III-A details the similarity mapping method from diffusion policy outputs to DMP parameter estimation, and Section III-B introduces the formulation of DMPs under surface constraints that are based on geodesic length.

\subsection{Surface-Constrained Motion Mapping Method}

In each inference cycle, the diffusion policy outputs a sequence of discrete desired actions
\begin{equation}
\mathcal{A}_{t} = \left\{ \left( p_{t}^{\mathrm{pred}}, q_{t}^{\mathrm{pred}} \right) \right\}_{t=t_{0}}^{t_{0}+N}
\end{equation}
which includes the end-effector positions and orientations over $N$ future time steps. During surface mapping, only the position component $p_{t}^{\mathrm{pred}}$ is projected onto the task surface, whereas the corresponding orientation $q_{t}^{\mathrm{pred}}$ is estimated afterward using an intrinsic orientation clustering approach.

To preserve the structural characteristics of the original trajectory after projection onto the task surface $S(u,v)$, the mapping process is formulated as a joint optimization problem. The discrete trajectory points that are generated by the diffusion policy are denoted as $\{ \hat{P}_i \}_{i=1}^{N}$; the goal is to find a corresponding set of surface points $\{ P_i^{\mathrm{surf}} = S(u_i, v_i) \}_{i=1}^{N}$ such that the projected points remain close to the original points in Euclidean space and the overall trajectory structure is retained. To obtain the surface-constrained position trajectory ${P_i^{\mathrm{surf}}}_{i=1}^N$, we minimize the following joint loss function:
\begin{equation}
\mathcal{L}_{\mathrm{total}} = \sum_{i=1}^{N} \left\| P_i^{\mathrm{surf}} - \hat{P}_i \right\|^2 + \lambda \sum_{i=2}^{N-1} D_{\mathrm{KL}} \left( \mathcal{H}_i^{\mathrm{surf}} \| \mathcal{H}_i^{\mathrm{raw}} \right)
\label{eq:joint_loss}
\end{equation}
where the first term enforces geometric consistency by penalizing the Euclidean distance between projected and original points and the second term evaluates structural similarity by measuring the Kullback--Leibler (KL) divergence between the local direction entropies before and after projection. $\mathcal{H}_i^{\mathrm{surf}}$ and $\mathcal{H}_i^{\mathrm{raw}}$ denote the first-order directional entropies of the $i$-th point in the projected and original trajectories, respectively. The KL divergence term explicitly constrains local structure distortion and prevents excessive smoothing.

The directional entropy $\mathcal{H}$ is computed as follows:
\begin{equation}
\mathcal{H} = -\sum_{j=1}^{J} P^{(j)} \log P^{(j)}
\end{equation}
where
\begin{equation}
P^{(j)} = \frac{1}{N-1} \sum_{i=1}^{N-1} \mathbb{I} \left( \frac{\Delta p_i}{\| \Delta p_i \|} \in \text{bin}_j \right)
\end{equation}
indicates the proportion of the differential vectors that fall into the $j$-th direction bin. By minimizing the KL divergence between the pre- and postmapping direction entropy distributions, the mapping process preserves the local geometric consistency of the trajectory.

After the surface-projected trajectory is obtained, the corresponding orientation trajectory can be estimated using Gaussian intrinsic clustering. To estimate the desired orientation $q_p$ at any target point $p_p$, we use the set of $K$ neighboring demonstration orientations $\{ q_i \}_{i=1}^{K}$. The orientation clustering center is obtained by minimizing the squared intrinsic distance on the unit quaternion manifold $\mathcal{S}^3$, which is defined as follows:
\begin{equation}
\underset{q_D}{\arg\min} \sum_{i=1}^{K} \left\| 2 \log^q \left( q_i * \overline{q_D} \right) \right\|^2
\end{equation}
where $\log$ denotes the logarithmic map and $*$ denotes quaternion multiplication.

While this approach performs well under uniformly distributed demonstrations, real-world data are typically nonuniformly distributed, which may lead to inaccurate orientation estimation. Therefore, we introduce a spatial weighting factor that is based on the distance between $p_D$ and each $p_i$, which yields the following weighted optimization objective:
\begin{equation}
\underset{q_D}{\arg\min} \sum_{i=1}^{K} \frac{d - \| p_D - p_i \|}{d} \cdot \left\| 2 \log^q \left( q_i * \overline{q_D} \right) \right\|^2
\label{eq:weighted_orientation}
\end{equation}
where $d$ represents a normalization constant.

\subsection{Robot Dynamic Motion Primitives}
DMPs encode robot motions as stable and convergent nonlinear dynamical systems, and offer good trajectory adaptability, disturbance rejection, and generalization capabilities. Therefore, the DMP framework is adopted in this work to generate robotic trajectories. To eliminate the influence of velocity profiles on spatial trajectory shapes, arc-length-based Dynamic Movement Primitives (AL-DMPs), following the formulation proposed by Gašpar et al. \cite{Gašpar2018}, are adopted for position trajectory encoding. Similar phase-independent formulations have also been studied in the literature, such as the work of Braglia et al.~\cite{Braglia2025}., which decouples the geometric path from the timing law. In this work, AL-DMPs are used as a trajectory representation to ensure that the spatial shape of the generated trajectories remains invariant under different execution speeds, while geodesic Dynamic Movement Primitives (Geo-DMPs) \cite{ke2025} are employed for orientation trajectory generation. In the proposed framework, both AL-DMPs and Geo-DMPs use geometric length (arc length or geodesic length) rather than time as the phase variable, thereby decoupling spatial trajectory shapes from temporal execution profiles and ensuring geometric consistency of the trajectories under different execution speeds.

AL-DMPs transform temporal derivatives into arc-length derivatives, thereby decoupling temporal and spatial evolution. The AL-DMP system is formulated as follows:
\begin{align}
  L\mathbf{z}' &= \alpha_z(\beta_z(g - y) - \mathbf{z}) + \mathbf{F}(x) \\
  L\mathbf{y}' &= \mathbf{z} \\
  Lx' &= -\alpha_x x
\end{align}
where $\mathbf{y} \in \mathbb{R}^3$ denotes the current end-effector position, $\mathbf{z} \in \mathbb{R}^3$ represents the velocity state, and $\mathbf{g} \in \mathbb{R}^3$ represents the target position. The variable $x \in [0,1]$ is the phase variable, $L$ represents the total arc length of the trajectory, and $\alpha_z$ and $\beta_z$ represent the damping and stiffness parameters, respectively. $\alpha_x$ controls the decay rate of the phase variable. $\mathbf{F}(x)$ represents the nonlinear forcing term, which is typically implemented as a weighted sum of radial basis functions to characterize the trajectory shape.

In contrast, the Euclidean formulation of DMPs cannot be directly applied to orientations that are represented as quaternions in $\mathbb{S}^3$. To address this, we employ the Geo-DMPs proposed in our previous study, which define rotational evolution along geodesics on the unit quaternion manifold. Unlike conventional time-based orientation DMP formulations, Geo-DMPs explicitly exclude the time variable and instead use geodesic length as the phase variable. This property allows Geo-DMPs to be naturally synchronized with AL-DMPs in the arc-length domain, thereby enabling consistent position–orientation coordination under variable execution speeds. The Geo-DMP system is expressed as follows:
\begin{align}
  L\boldsymbol{\eta}' &= \alpha_z \left( \beta_z \cdot 2\log^q(q_g * \bar{q}) - \boldsymbol{\eta} \right) + \mathbf{f}_q(x) \\
  L\mathbf{q}' &= \frac{1}{2} \boldsymbol{\eta} * \mathbf{q} \\
  Lx' &= -\alpha_x x
\end{align}
where $\mathbf{q} \in \mathbb{S}^3$ represents the current orientation quaternion and $q_g \in \mathbb{S}^3$ represents the target orientation. The variable $\boldsymbol{\eta} \in \mathbb{R}^3$ represents the derivative of the orientation along the geodesic length $L$, which is treated as a purely imaginary quaternion. $\log^q(\cdot)$ denotes the logarithmic map that is used to measure the shortest paths on the sphere, and $*$ represents quaternion multiplication. The nonlinear term $\mathbf{f}_q(x)$ encodes the orientation trajectory shape, and the other parameters share the same meanings as in AL-DMP.

The geodesic length at time step $n$ is computed as follows:
\begin{equation}
g_n = 
\begin{cases}
  \sum\limits_{k=1}^{n} 2\left\| \log^q\left( \mathbf{q}_k * \overline{\mathbf{q}_{k+1}} \right) \right\|, & n \ge 2 \\
  0, & n = 1
\end{cases}
\end{equation}

Given the initial condition $x(0) = 1$, the phase variable can be analytically solved from \eqref{eq:weighted_orientation} as follows:
\begin{equation}
x = x(g_n) = \exp\left( -\frac{\alpha_x}{L} g_n \right).
\label{eq:phase}
\end{equation}

This solution demonstrates that the phase variable decays exponentially with the geodesic distance, which enables the system to generate rotational motions that are both smooth and physically consistent in the orientation space. Moreover, as described in \eqref{eq:phase}, AL-DMPs and Geo-DMPs can be phase-synchronized through the geodesic-length term, ensuring consistent spatial correspondence between position and orientation trajectories.

\begin{figure*}[!t]
\centering
\includegraphics[scale=0.41]{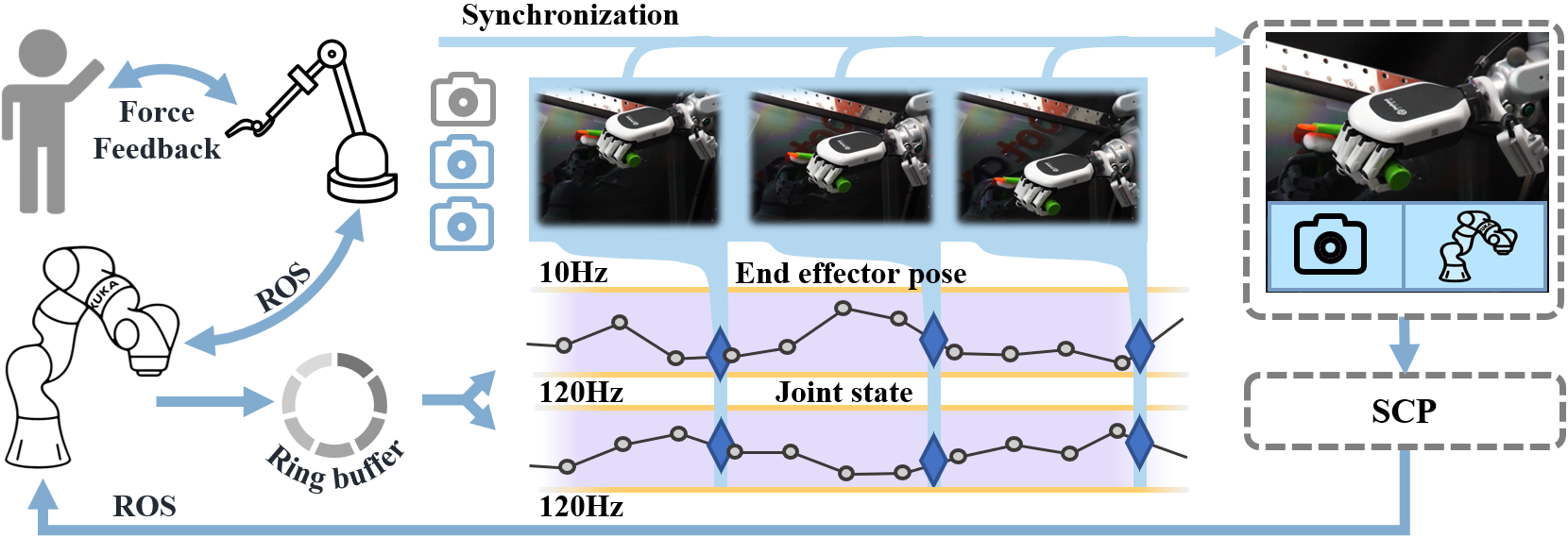}
\caption{Illustration of the teleoperation and data acquisition architecture. The system is built on an ROS for robot control and data flow management, which supports force--feedback interactions, the synchronized acquisition of camera images and robot states, and the feeding of aligned data to the SCP module via a ring buffer. The control loops of the robot and the haptic device run at 160 Hz, whereas the pose and joint states are sampled at 120 Hz to ensure synchronization with the visual data stream.}
\label{fig4}
\end{figure*}

\section{Experiment}
To validate the effectiveness of the proposed approach, we designed three tasks with progressively increasing task complexity to evaluate the performance of the SCP, namely whiteboard wiping, free-form surface wiping, and aircraft windscreen cleaning. In these tasks, we conducted comparative experiments focusing on task success rate, completion efficiency, surface-alignment error (SAE), motion acceleration, and trajectory curvature. Several state-of-the-art methods—DP~\cite{chi2023}, MDP~\cite{scheikl2024}, and ACP~\cite{hou2024}—were selected as the baselines for comparison. DP represents a classical diffusion-based policy that does not introduce any explicit geometric or dynamic constraints. MDP improves trajectory smoothness and dynamic feasibility by combining diffusion models with movement primitives. ACP is a contact-force-regulation-based strategy that maintains stable contact through impedance control.

To collect human demonstration data, we developed a teleoperation platform with a force-feedback capability. The platform consisted of a KUKA iiwa14 robot and a Haption 6D haptic device. The KUKA iiwa14 robot estimated end-effector contact forces through its joint torque sensors, and the measured forces were fed back to the operator via the Haption device. The behaviors of the robot were recorded using three Intel D435i cameras, namely, one eye-in-hand camera and two fixed cameras.

The SCP used an observation horizon $T_o = 2$, an action horizon $T_a = 8$, and a prediction horizon $T_p = 16$. The image inputs were randomly cropped to a size of $3 \times 230 \times 230$. The diffusion network adopted a 1D convolutional architecture. During training, the batch size was set to 64, and the Adam optimizer was used with a learning rate of $1 \times 10^{-4}$ and a weight decay of $1 \times 10^{-6}$. Cosine learning rate decay was applied.
For both training and inference, a 100-step DDPM process was used, whereas in the real-world robot experiments, a 16-step DDIM was used to improve the inference speed.

In the following sections, we provide an overview of each task, the evaluation metrics that were used for each task, and the corresponding experimental results.

\begin{figure}[t]
    \centering
    \includegraphics[width=0.4\textwidth]{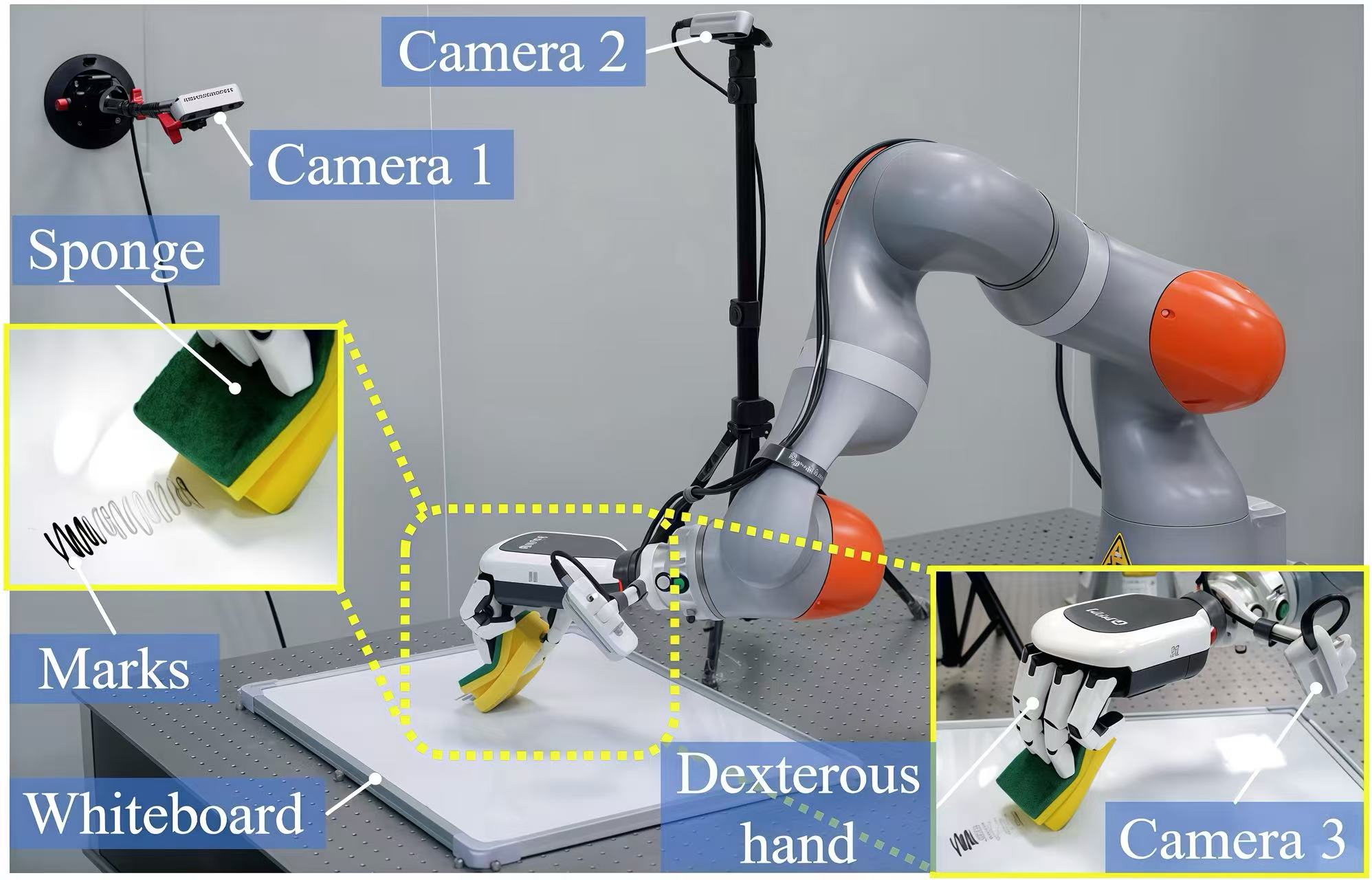}
    \caption{Whiteboard wiping experiment setup.}
    \label{fig5}
\end{figure}

\begin{table}[t]
    \centering
    \caption{Performance comparison in the whiteboard wiping task.}
    \label{tab:whiteboard}
    \vspace{1pt}
    \scalebox{0.9}{
    \begin{tabular}{l|c|cccc}
        \toprule
        & Human & DP & MDP & ACP & \textbf{SCP} \\
        \midrule
        Succ $\uparrow$ & 1.00 & 0.59$\pm$0.03 & 0.92$\pm$0.02 & 0.99$\pm$0.02 & \textbf{0.99$\pm$0.01} \\
        AWT $\downarrow$ & 1.75 & 3.22$\pm$0.26 & 2.53$\pm$0.18 & 1.89$\pm$0.09 & \textbf{1.82$\pm$0.10} \\
        \bottomrule
    \end{tabular}
    }
\end{table}

\subsection{Whiteboard Wiping}
The whiteboard wiping task represents a relatively simple experimental scenario with weak geometric constraints. In this setting, task success mainly depends on the robot’s ability to generate smooth, continuous, and dynamically feasible motion trajectories. the robot is required to remove marker traces of random shapes and positions along the surface of a whiteboard, and the task is considered successful only when all marks are completely erased. The robot’s end effector, a dexterous hand, grasps a sponge as the wiping tool, as illustrated in Fig. \ref{fig5}. We collected 200 teleoperated demonstrations of different stain distributions, where the human wiping behaviors exhibited diverse motion patterns depending on the shape and location of the marks (see the supplementary video). For each method, we conducted 5 groups of 20 trials, and the success rate (Succ) as well as the average number of wipes (AWT, counted only for successful trials) were evaluated, as summarized in Table~\ref{tab:whiteboard}.

As the most basic diffusion-based method, DP produced significantly lower success rates than SCP and ACP, and it also showed a clear disadvantage in the average number of wipes. The inherent reason is that DP generates actions with substantial randomness and lacks any mechanism to align the inferred actions with the robot’s current state. As illustrated by the acceleration profiles and trajectory curvature in Fig. \ref{fig：SAE}, such misalignment often leads to abrupt acceleration spikes and curvature discontinuities, causing momentary loss of contact or instantaneous overload, which ultimately results in task failure.

\begin{figure}[t]
    \centering
    \includegraphics[width=0.40\textwidth]{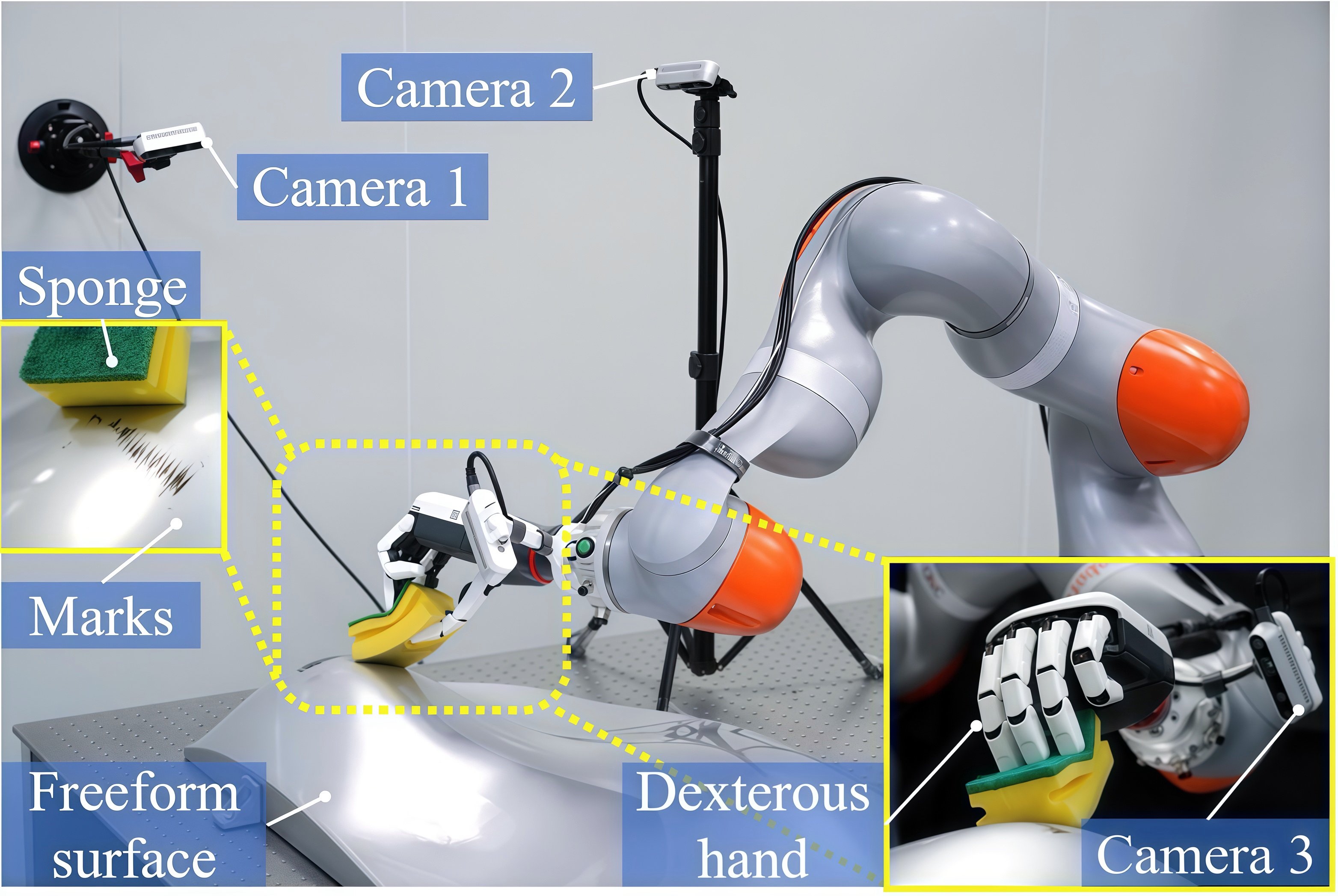}
    \caption{Free-form surface wiping experimental setup.}
    \label{fig8}
\end{figure}

\begin{table}[t]
    \centering
    \caption{Performance comparison in the free-form surface wiping task.}
    \label{tab:freeform}
    \vspace{1pt}
    \scalebox{0.9}{
    \begin{tabular}{l|c|cccc}
        \toprule
        & Human & DP & MDP & ACP & \textbf{SCP} \\
        \midrule
        Succ $\uparrow$ & 1.00 & -- & 0.52$\pm$0.02 & 0.97$\pm$0.03 & \textbf{0.98$\pm$0.02} \\
        AWT $\downarrow$ & 1.35 & -- & 2.51$\pm$0.20 & 1.50$\pm$0.08 & \textbf{1.42$\pm$0.09} \\
        \bottomrule
    \end{tabular}
    }
\end{table}

The MDP effectively improves motion continuity, as actions generated by its DMP formulation are smoother, exhibiting fewer sudden accelerations and more regular curvature patterns Fig. \ref{fig:acc}. This prevents collisions or unintended separation from the surface. However, because MDP still lacks explicit encoding of planar geometric constraints, it cannot ensure consistent adherence to the whiteboard surface. Consequently, its surface-alignment error remains relatively large Fig. \ref{fig：SAE}, and the success rate is still significantly lower than those of ACP and SCP.

The ACP maintains a stable normal contact force during wiping through impedance control, allowing the tool to remain consistently attached to the whiteboard surface. As shown in the surface-alignment error curves in Fig. \ref{fig：SAE}, ACP achieves small planar deviation errors, leading to performance comparable to that of SCP. In the 200 human demonstration episodes, the average number of wipes was 1.75, and both ACP and SCP achieved similar AWT values that closely matched human performance. Therefore, we concluded that in planar wiping tasks, both the contact-force-based ACP method and the geometry-constraint-based SCP method can effectively learn and reproduce human wiping skills.

\subsection{Free-Form Surface Wiping}
The geometric characteristics of free-form surfaces are highly complex and vary significantly across different regions, which places more stringent requirements on motion generation and real-time adaptation. In addition to generating smooth motions, the tool trajectory is required to closely conform to the target surface geometry, as loss of contact will lead to task failure. The experiment setup is illustrated in Fig. \ref{fig5}. Similar to the planar wiping task, the operation is considered successful only when all marker traces on the surface are completely removed. Across all 200 teleoperated demonstrations, the human operator achieved an average wipe count of 1.35. The experimental results of the proposed method and the baselines are summarized in Table~\ref{tab:freeform}, where DP results are omitted due to frequent task failures and insufficient valid executions.

\begin{figure*}[!t]
\centering
\subfloat{\includegraphics[scale=0.3]{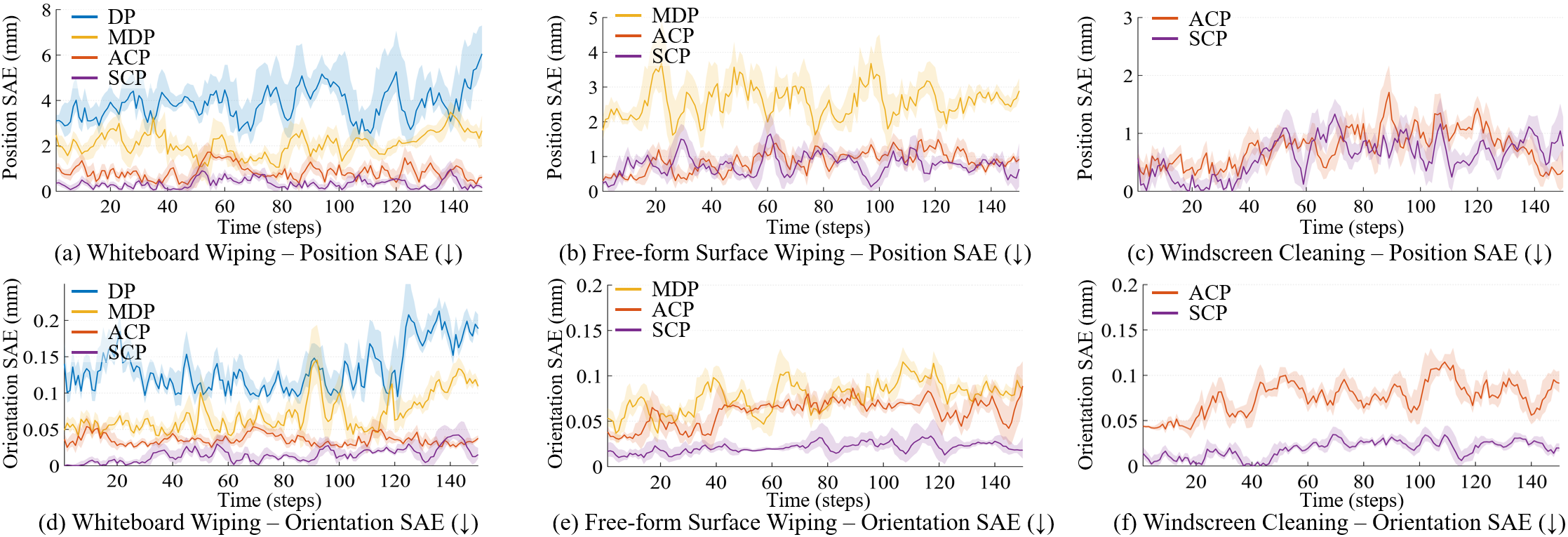}}%
\hfil
\caption{Comparison of position and orientation surface-alignment error (SAE) across the three tasks. For each task, each curve was computed using 10 trajectories, obtained by selecting two similar wiping trajectories from each of five independent experimental groups, where the stain locations and shapes were kept consistent within each group. The shaded regions denote the standard deviation. Figures (a)–(c) show the position surface-alignment error for the whiteboard wiping, free-form surface wiping, and windscreen cleaning tasks, respectively, while (d)–(f) present the corresponding orientation surface-alignment error.}
\label{fig：SAE}
\end{figure*}

The success rate of MDP was substantially lower than those of SCP and ACP. Although MDP ensures motion continuity through its DMP structure, it does not explicitly encode or define the robot’s action primitives. As a result, all task-related skills are implicitly embedded within the policy, and the generated motions lack task-relevant surface geometry constraints. Consequently, as illustrated in Fig. \ref{fig：SAE}, MDP exhibits significantly higher surface-alignment error on free-form surfaces compared to ACP and SCP. In practice, this often leads to two types of failure behaviors: momentary loss of contact with the surface or excessive compression of the sponge, both of which negatively impact the wiping outcome.

\begin{figure}[t]
    \centering
    \includegraphics[width=0.389\textwidth]{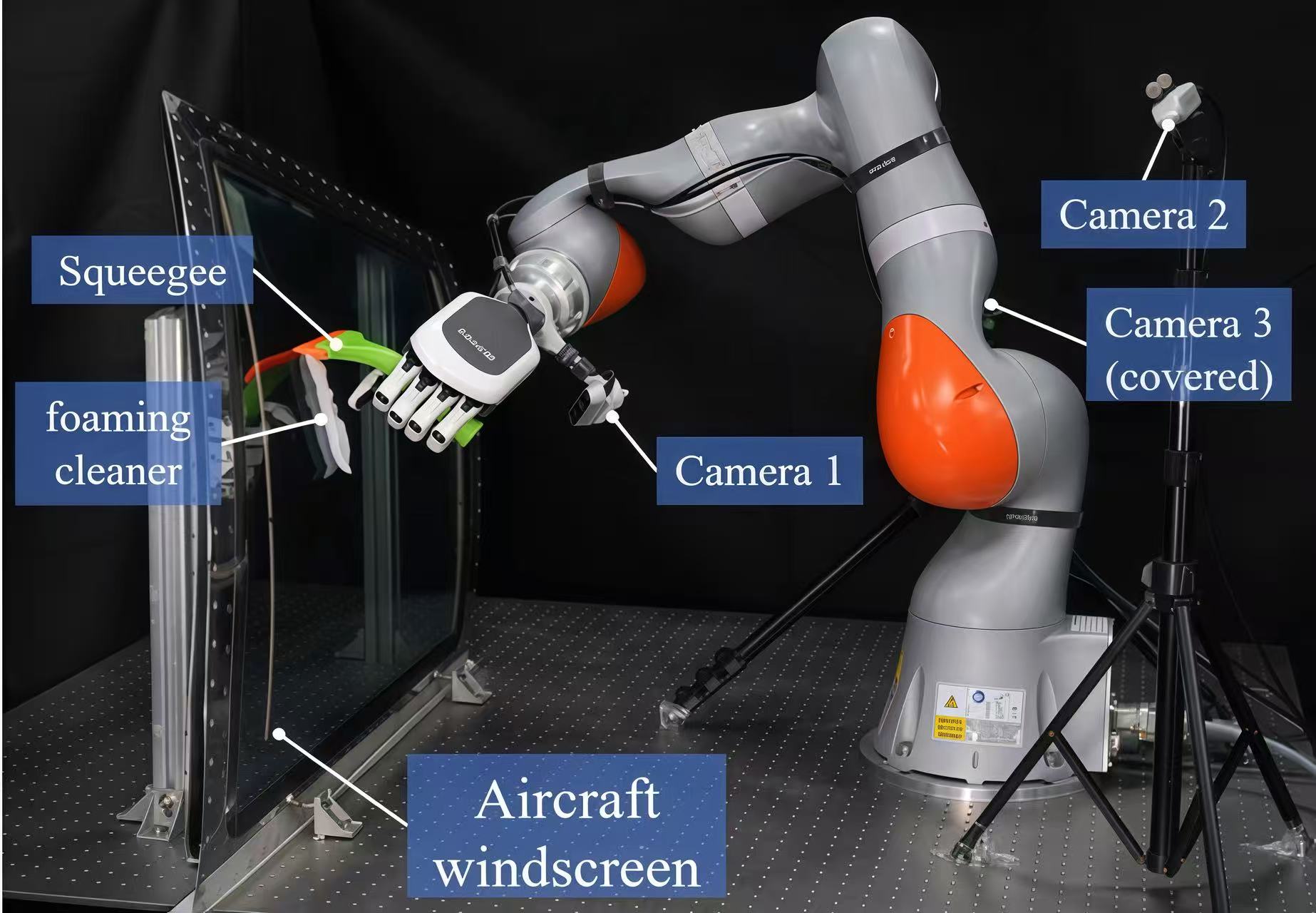}
    \caption{Windscreen cleaning experiment setup.}
    \label{fig9}
\end{figure}

\begin{table}[t]
    \centering
    \caption{Performance comparison in the windscreen cleaning task.}
    \label{tab:windscreen}
    \vspace{1pt}
    \scalebox{1.05}{
    \begin{tabular}{l|c|cccc}
        \toprule
        & Human & DP & MDP & ACP & \textbf{SCP} \\
        \midrule
        Succ $\uparrow$ & 1.00 & -- & -- & 0.63$\pm$0.02 & \textbf{0.98$\pm$0.02} \\
        \bottomrule
    \end{tabular}
    }
\end{table}

In contrast, ACP and SCP both achieved performance levels comparable to human demonstrations. ACP relies on impedance control to regulate contact force, thereby maintaining stable attachment of the sponge to the curved surface. SCP, on the other hand, enforces explicit surface geometry constraints that ensure the wiping trajectory consistently adheres to the underlying surface shape. It is noteworthy that in this task scenario, the sponge used by the robot inherently tolerates relatively large deviations in contact position and orientation, resulting in lower precision requirements for motion generation. Thus, even though ACP and SCP exhibit noticeable differences in surface-alignment error (Fig. \ref{fig：SAE}), their success rates remain similarly high.

\subsection{Aircraft Windscreen Cleaning}

Aircraft maintenance personnel typically use a squeegee to clean the windscreen, as shown in Fig.~\ref{fig9}. Compared with the sponge used in the previous two experiments, the squeegee is significantly more rigid and requires much higher contact precision. Once contact is established, the squeegee must maintain an appropriate contact angle and uniformly adhere to the glass surface throughout the motion; otherwise, incomplete cleaning or even equipment damage may occur. We constructed an experimental platform in which the operator sprays foaming cleaner onto the aircraft windscreen, and the robot grasps a squeegee to perform the cleaning task. The selected windscreen is an irregular free-form surface, and the operator applies the foaming cleaner at arbitrary positions and shapes across the glass.

\begin{figure*}[!t]
\centering
\subfloat{\includegraphics[scale=0.3]{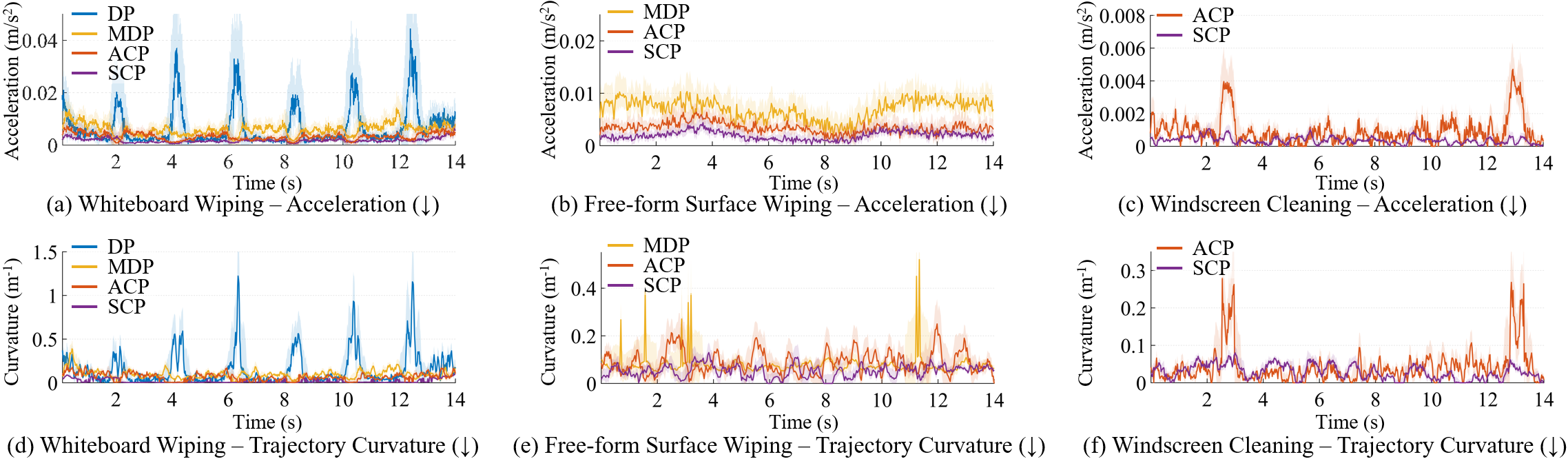}}%
\hfil
\caption{Comparison of acceleration profiles and trajectory curvature across the three tasks. For each task, each curve was computed using 10 trajectories, obtained by selecting two similar wiping trajectories from each of five independent experimental groups, where the stain locations and shapes were kept consistent within each group. The shaded regions denote the standard deviation. Figures (a)–(c) show the acceleration profiles for the whiteboard wiping, free-form surface wiping, and windscreen cleaning tasks, respectively, while (d)–(f) present the corresponding trajectory curvature.}
\label{fig:acc}
\end{figure*}

\begin{figure}[t]
    \centering
    \includegraphics[scale=0.78, angle=0, origin=c]{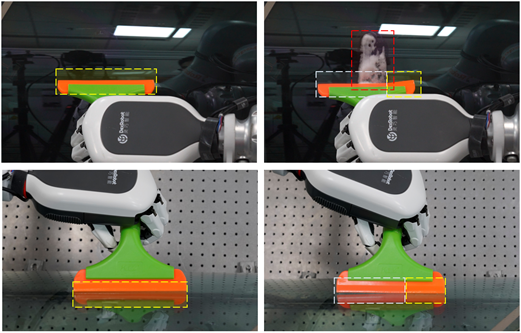}
    \caption{Contact status of the squeegee and corresponding wiping performance. The regions that are outlined by yellow dashed boxes are areas where the squeegee blade maintained firm contact with the glass surface. The areas that are outlined by light blue dashed boxes are regions where contact was lost. In the regions that are outlined by red dashed boxes, residual foaming cleaner was left after wiping.}
    \label{fig10}
\end{figure}

In the windscreen cleaning task, neither DP nor MDP consistently completed the task, and occasional unsafe behaviors posing a risk to the experimental tool were observed. Therefore, their results are not reported in Table~\ref{tab:windscreen}, and ACP was selected as the baseline method for comparison. The ACP achieved a success rate of 0.63 ± 0.02. During the experiments, although ACP was able to leverage contact force feedback to maintain consistent contact between the squeegee and the glass, and the robot’s impedance controller provided a degree of posture compliance, occasional posture errors still occurred. These errors caused one-sided lifting or angular deviation of the squeegee during motion. As shown in Fig. \ref{fig：SAE}, the posture surface-alignment error of ACP is significantly higher than that of SCP, and improper squeegee orientation often prevented the blade edge from adhering uniformly to the glass, resulting in incomplete cleaning, as illustrated in Fig.~\ref{fig10}.

In contrast, the SCP method integrates the geometric constraints encoded by surface DMPs with the multistep refinement capability of the diffusion policy. This formulation not only ensures correct alignment between the squeegee and the surface’s normal–tangential directions during the initial trajectory planning but also introduces corrective adjustments at every diffusion inference step. During the execution of surface DMPs, SCP provides real-time compensation for position and orientation drifts caused by external disturbances or model inaccuracies. The experimental results show that SCP achieved a 0.98 ± 0.02 success rate, approaching the performance of human operators. Moreover, SCP demonstrated strong generalization ability and behavioral diversity. These findings indicate that SCP effectively improves the robustness, precision, and dynamic feasibility of robots performing cleaning tasks on high-curvature surfaces and provides a promising approach for skill learning in complex surface wiping scenarios.

\subsection{Discussion}
SCP outperforms the baseline methods in the above experiments. Compared with existing approaches, SCP fundamentally avoids two common failure modes in complex surface manipulation tasks: inaccurate geometric intention inference and unstable action execution. 

On the one hand, SCP explicitly incorporates surface geometric constraints into the policy inference stage through geometric encoding, which directly explains its significant advantage in terms of Surface-Alignment Error (SAE) in free-form surface tasks. In contrast, DP and MDP mainly rely on implicitly learning geometric relationships from data distributions, which leads to inferior SAE performance. On the other hand, SCP converts policy intentions into dynamically feasible and geometrically consistent trajectories through a similarity-based mapping mechanism. This mechanism effectively suppresses abrupt action variations, thereby explaining SCP’s overall advantages in motion acceleration, trajectory curvature, and task success rate.

\section{Conclusions}
In this paper, a robotic action generation strategy for surface-constrained tasks, which is termed the SCP, is presented. The proposed method integrates a conditional diffusion model with surface-aware DMPs to enable the generation of robot actions that comply with free-form surface constraints on the basis of human demonstrations and real-time visual observations. It achieves a unified formulation of structured surface-geometry intent and dynamically feasible action execution.

In experiments on three representative scenarios, namely, whiteboard wiping, complex free-form surface cleaning, and aircraft windshield polishing, the proposed method consistently outperformed several baseline methods (DP, MDP, and ACP) in terms of task success rate, trajectory smoothness, and contact stability. Notably, a perfect success rate of 100\% was achieved across all the tasks, thus validating the effectiveness of the proposed strategy in surface-constrained robotic applications.

Despite these promising results, the current method depends heavily on high-quality demonstration data. Future research will focus on incorporating prior knowledge from large language models and video-language models to enable rapid adaptation and generalization to novel geometries, complex tools, and previously unseen tasks with only a few demonstrations. Such advancements would further increase the practicality and intelligence of the framework in complex industrial contact-rich scenarios.

\vspace{-0.3in}
\begin{IEEEbiography}[{\includegraphics[width=1.05in,height=1.25in,clip,keepaspectratio]{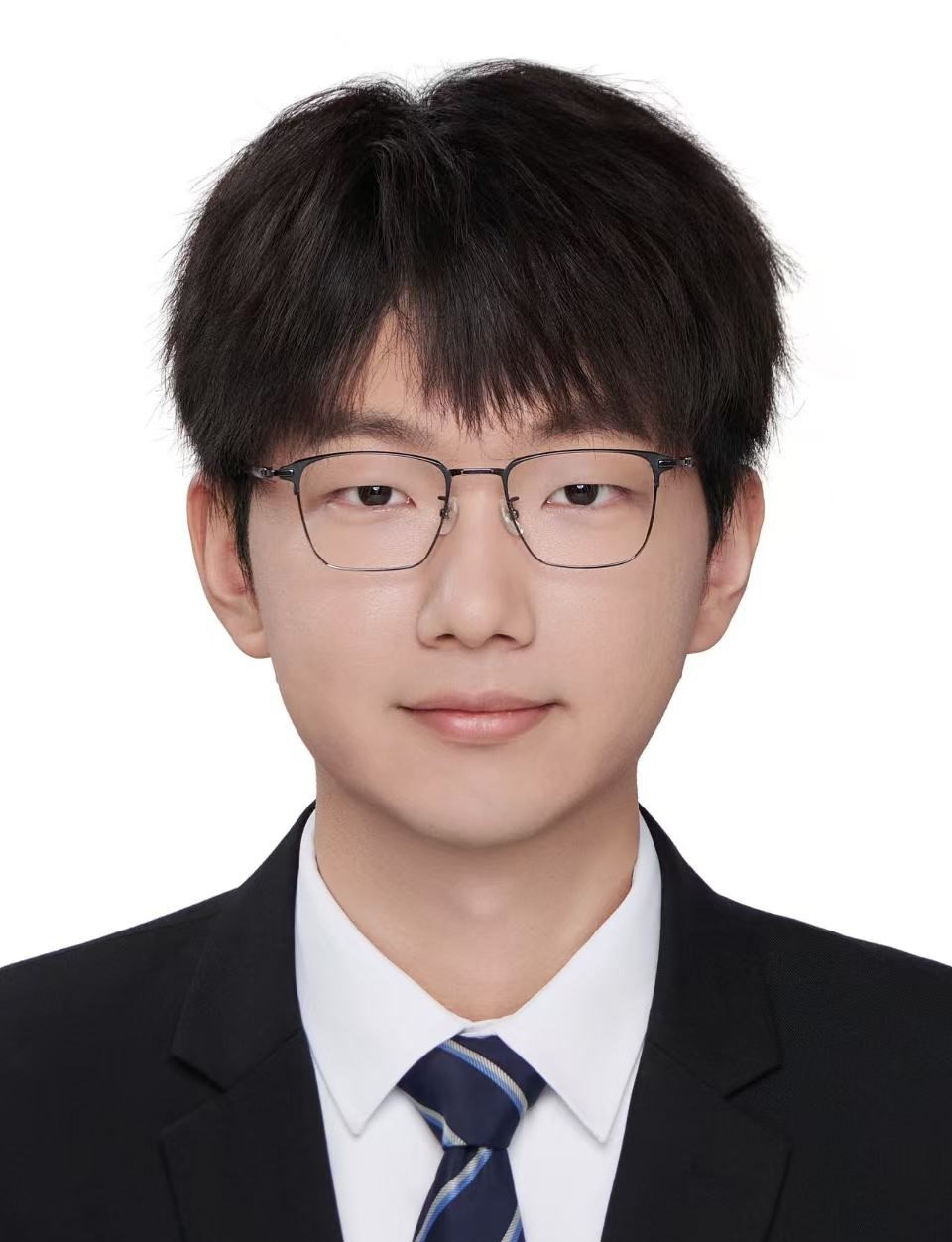}}]{Shuai Ke}
received the B.E. degree from the School of Automation, China University of Geosciences, Wuhan, China, in 2023. He is pursuing a Ph.D. in mechanical engineering at Huazhong University of Science and Technology, Wuhan, China. His research interests include embodied dexterous manipulation and robot imitation learning.
\end{IEEEbiography}
\vspace{-0.38in}
\begin{IEEEbiography}[{\includegraphics[width=1in,height=1.25in,clip,keepaspectratio]{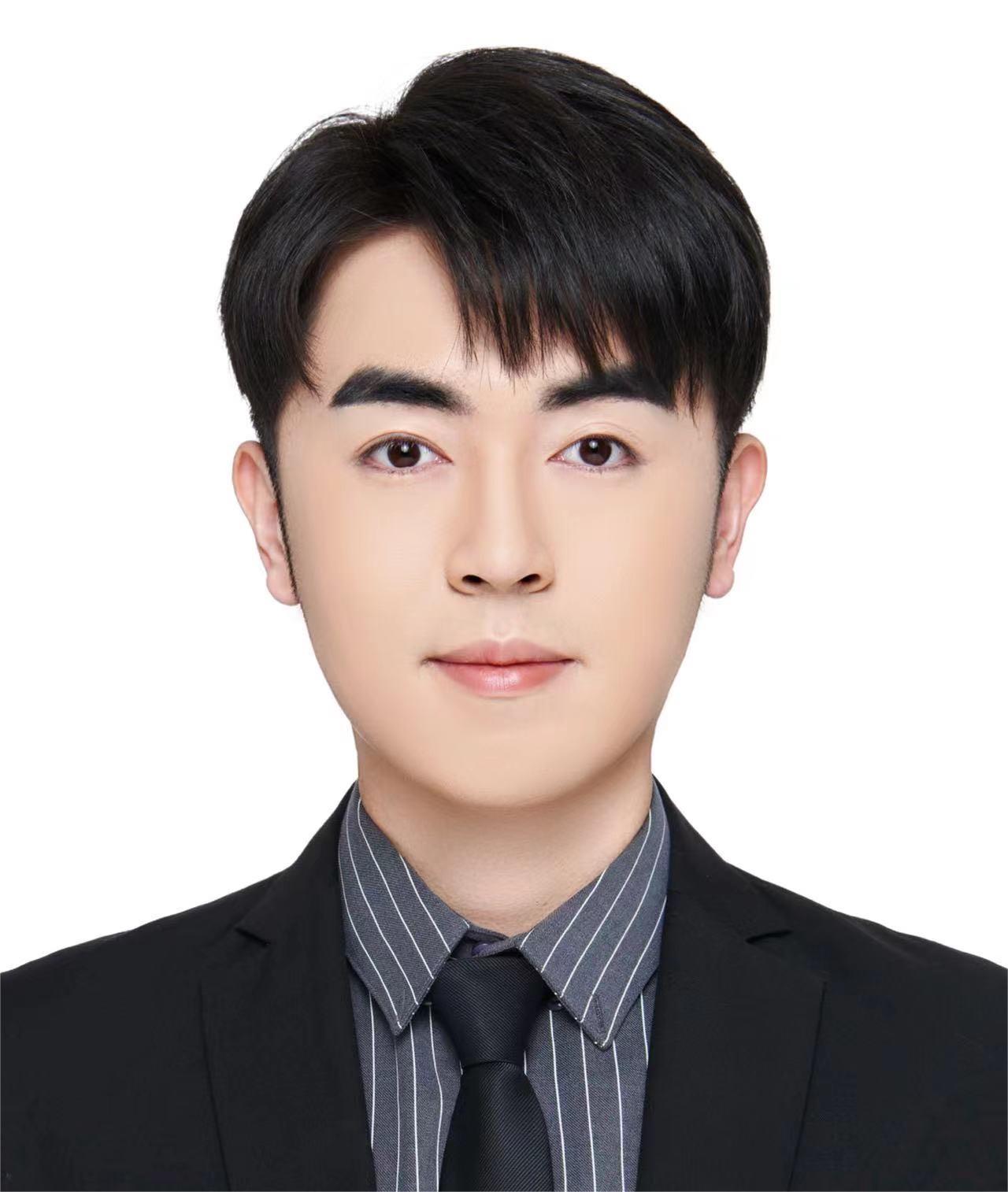}}]{Jiexin Zhang}
 (Member, IEEE) was born in Shandong, China, and received his B.E. degree and Ph.D. from Shandong University and Shanghai Jiao Tong University in 2018 and 2024, respectively. 
He is currently an assistant researcher at the State Key Laboratory of Intelligent Manufacturing Equipment and Technology, Huazhong University of Science and Technology. His current research interests include robotic dexterous manipulation, elastic system modeling, and compliance control.
\end{IEEEbiography}
\vspace{-0.3in}
\begin{IEEEbiography}[{\includegraphics[width=0.94in,height=1.25in,clip,keepaspectratio]{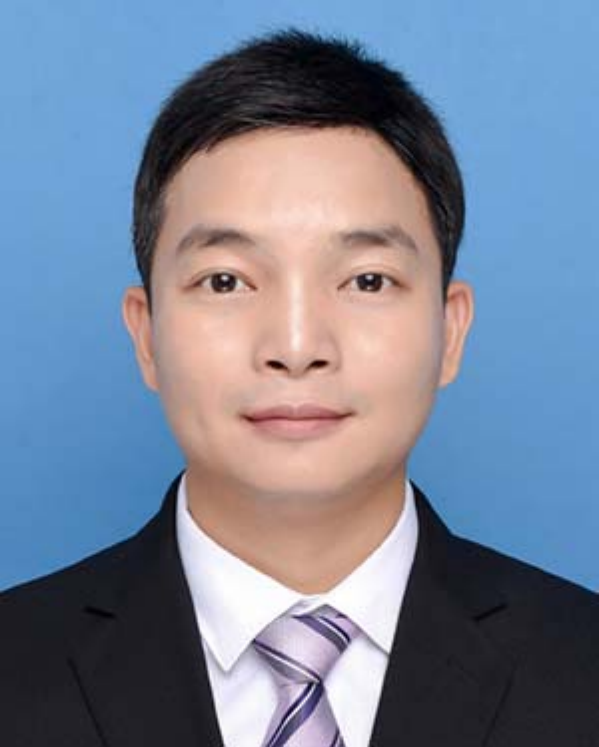}}]{Huan Zhao}
(Member, IEEE) received the B.S.
degree in mechanical engineering from Jilin University, Changchun, China, in 2006, and
the Ph.D. degree in mechanical engineering from Shanghai Jiao Tong University, Shanghai,
China, in 2013.
He conducted postdoctoral research with the Huazhong University of Science and Technology, Wuhan, China, in 2013, where he is currently a Professor. His research interests include robotic machining and assembly.
\end{IEEEbiography}
\vspace{-0.4in}
\begin{IEEEbiography}[{\includegraphics[width=1in,height=1.25in,clip,keepaspectratio]{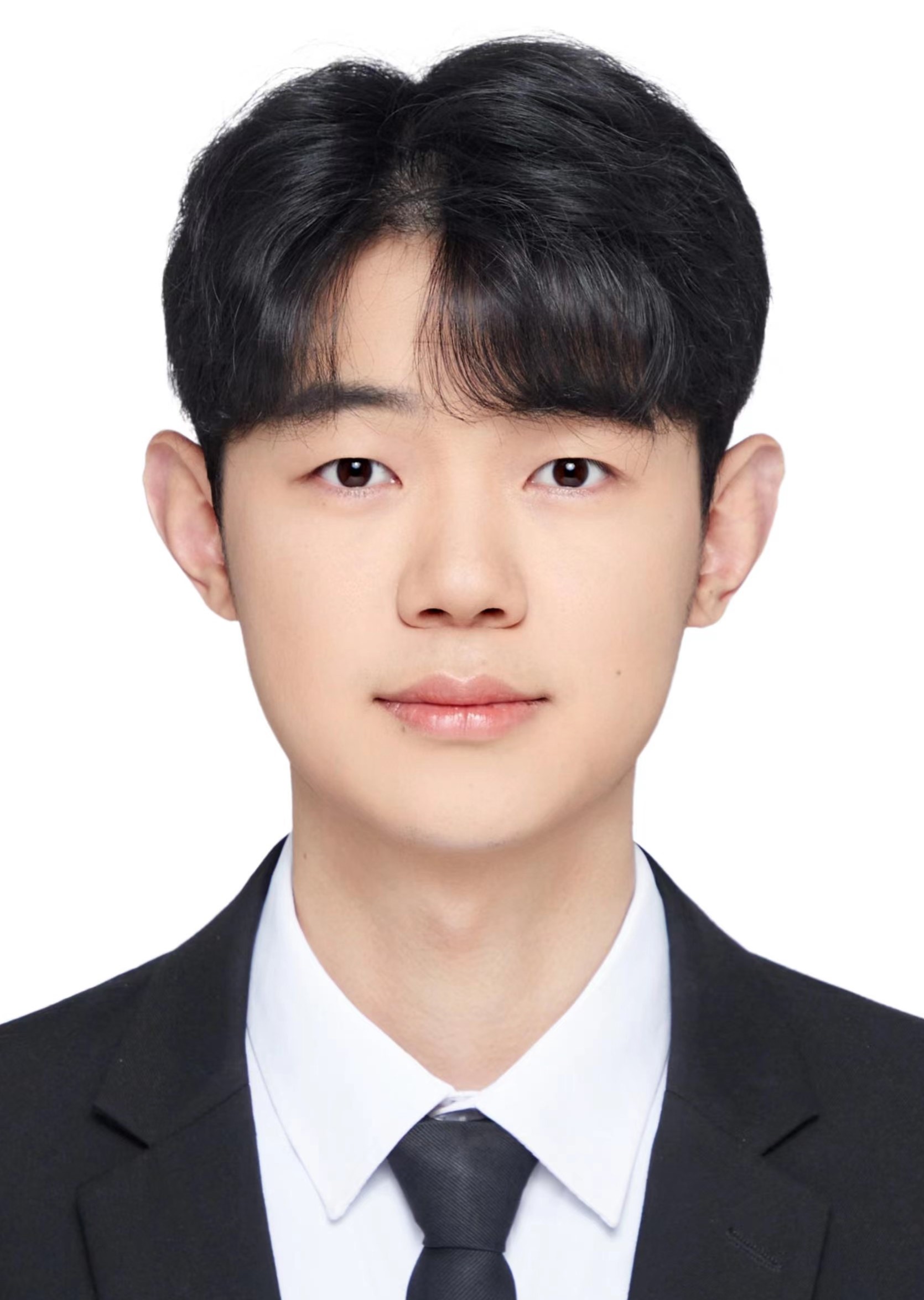}}]{Zhiao Wei}
received the B.E. degree in mechanical engineering from Nanjing University of Aeronautics and Astronautics, Nanjing, China, in 2023. He is currently pursuing an M.S. degree in mechanical engineering at Huazhong University of Science and Technology, Wuhan, China. His research interests include continual learning for robotic systems and embodied dexterous manipulation.
\end{IEEEbiography}
\vspace{-0.3in}
\begin{IEEEbiography}[{\includegraphics[width=1.1in,height=1.25in,clip,keepaspectratio]{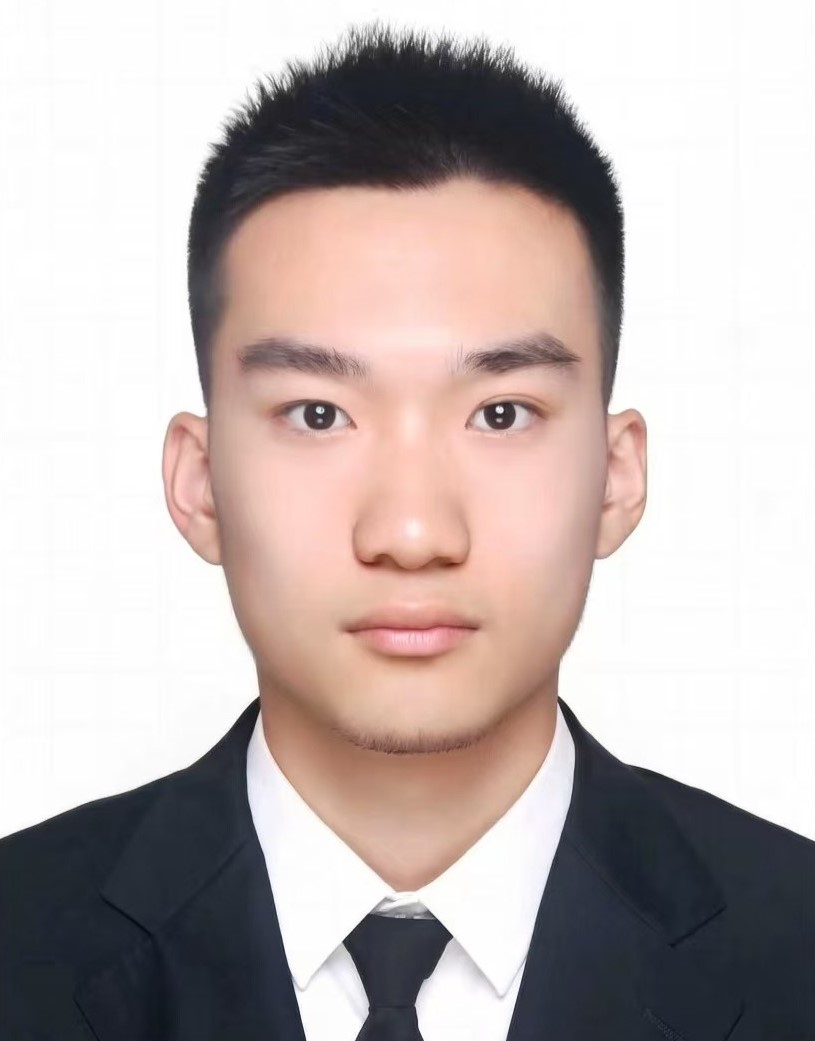}}]{Yikun Guo}
received the B.E.degree in Mechanical Design, Manufacturing, and Automation from Huazhong University of Science and Technology, Wuhan, China, in 2025.He is currently working toward the M.S. degree in mechanical engineering at Huazhong University of Science and Technology, Wuhan, China. His research interests include robot imitation learning, reinforcement learning, and robotic machining.
\end{IEEEbiography}
\vspace{-0.3in}
\begin{IEEEbiography}[{\includegraphics[width=1.1in,height=1.25in,clip,keepaspectratio]{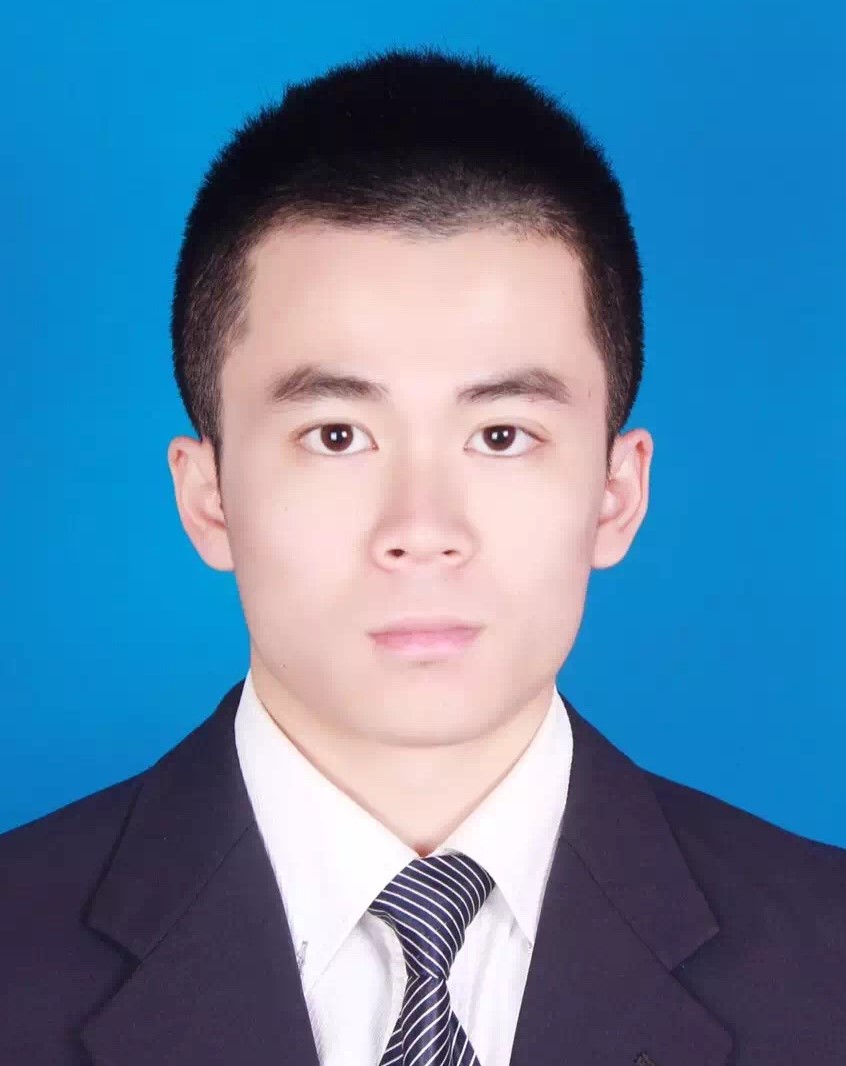}}]{Jie Pan}
received the M.E. degree from the School of Mechanical Engineering, Jiangsu University, Zhenjiang, China, in 2019. He is currently pursuing a Ph.D. degree with the School of Mechanical Science and Engineering, Huazhong University of Science and Technology, Wuhan, China. His research interests include robotic manufacturing and machine learning algorithms.
\end{IEEEbiography}
\vspace{-0.3in}
\begin{IEEEbiography}[{\includegraphics[width=1in,height=1.25in,clip,keepaspectratio]{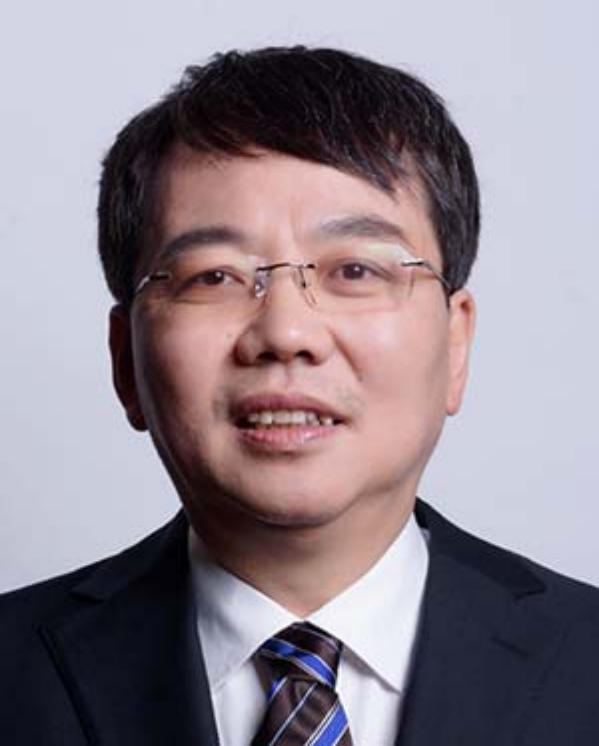}}]{Han Ding}
(Senior Member, IEEE) received the Ph.D. degree in mechanical engineering from the Huazhong University of Science and Technology (HUST), Wuhan, China, in 1989.
He has been a Professor with the HUST, since 1997. He was a Cheung Kong Chair Professor with Shanghai Jiao Tong University, Shanghai, China, from 2001 to 2006. His research interests include robotics, multiaxis machining, and control engineering.
Dr. Ding was elected a Member of the Chinese Academy of Sciences, in 2013.
\end{IEEEbiography}

\vfill

\end{document}